\pdfoutput=1

\documentclass[11pt]{article}

\usepackage[final]{acl}

\usepackage{gb4e} 
\noautomath

\usepackage{times}
\usepackage{latexsym}
\usepackage{amsfonts}
\usepackage{amsmath}
\usepackage{subcaption}
\usepackage{graphicx}
\usepackage{scalefnt} 
\usepackage{diagbox}

\usepackage{graphicx}
\usepackage{xcolor}
\usepackage{CJKutf8}
\usepackage{tabularx}
\usepackage{booktabs}
\usepackage{makecell}
\usepackage[misc]{ifsym}
\usepackage{amssymb}
\usepackage{pifont}
\usepackage[export]{adjustbox}
\usepackage{multirow}
\usepackage{langsci-gb4e}

\newcommand{\zhsmall}[1]{\begin{CJK*}{UTF8}{gbsn}\small{#1}\end{CJK*}}

\newcommand{\red}[1]{\textcolor{red}{#1}}

\usepackage[T1]{fontenc}

\usepackage[utf8]{inputenc}

\usepackage{microtype}

\usepackage{inconsolata}

\usepackage{graphicx}

\title{Read it in Two Steps: Translating Extremely Low-Resource Languages with Code-Augmented Grammar Books}

\author{Chen Zhang\thanks{Equal contribution.},\ \ Jiuheng Lin$^*$,\ \  Xiao Liu,\ \ Zekai Zhang,\ \  Yansong Feng\thanks{Corresponding author.} \\
Wangxuan Institute of Computer Technology, Peking University \\
{\tt \{zhangch,fengyansong\}@pku.edu.cn} \\
{\tt linjiuheng@stu.pku.edu.cn}}

\begin{document}
\maketitle
\begin{abstract}
While large language models (LLMs) have shown promise in translating extremely low-resource languages using resources like dictionaries, the effectiveness of grammar books remains debated. 
This paper investigates the role of grammar books in translating extremely low-resource languages by decomposing it into two key steps: grammar rule retrieval and application. 
To facilitate the study, we introduce \textsc{ZhuangRules}, a modularized dataset of grammar rules and their corresponding test sentences.
Our analysis reveals that rule retrieval constitutes a primary bottleneck in grammar-based translation.
Moreover, although LLMs can apply simple rules for translation when explicitly provided, they encounter difficulties in handling more complex rules.
To address these challenges, we propose to represent grammar rules as code functions, motivated by their similarities in structures and the benefit of code in facilitating LLM reasoning. 
Our experiments show that using code rules significantly boosts 
both rule retrieval and application, ultimately resulting in a 13.1\% BLEU improvement in translation.
\end{abstract}

\section{Introduction}

Most human languages suffer from data scarcity~\cite{joshi-etal-2020-state}.
With only a few thousand sentences available for extremely low-resource (XLR) languages, traditional pretraining or finetuning methods~\cite{yong-etal-2023-bloom, liu-etal-2020-multilingual-denoising} are impractical for building effective machine translation (MT) systems.
Facing the challenge of XLR MT, large language models (LLMs) offer a promising alternative. 
Recent research reveals that LLMs can perform XLR MT through in-context learning (ICL), leveraging small-scale linguistic resources like dictionaries and parallel sentences~\cite{tanzer2024a,zhang-etal-2024-teaching}. 
Among these resources, grammar books, with their systematic linguistic descriptions, appear ideal for guiding translation, but their effectiveness remains debated. 
Some studies claim that prompting LLMs with full grammar books improves translation performance~\cite{tanzer2024a, team2024gemini,zhang-etal-2024-hire}, while others argue that such improvements may stem from lexical leakage, where LLMs identify the bilingual explanations of several words in the test sentence from the grammar book and use them as shortcuts, rather than genuinely understanding grammar rules~\cite{aycock2024can}. 
However, no existing dataset effectively eliminates such interference factors, making it difficult to assess whether LLMs truly understand grammar rules.

\begin{figure}[t]
\centering
\includegraphics[scale=0.57]{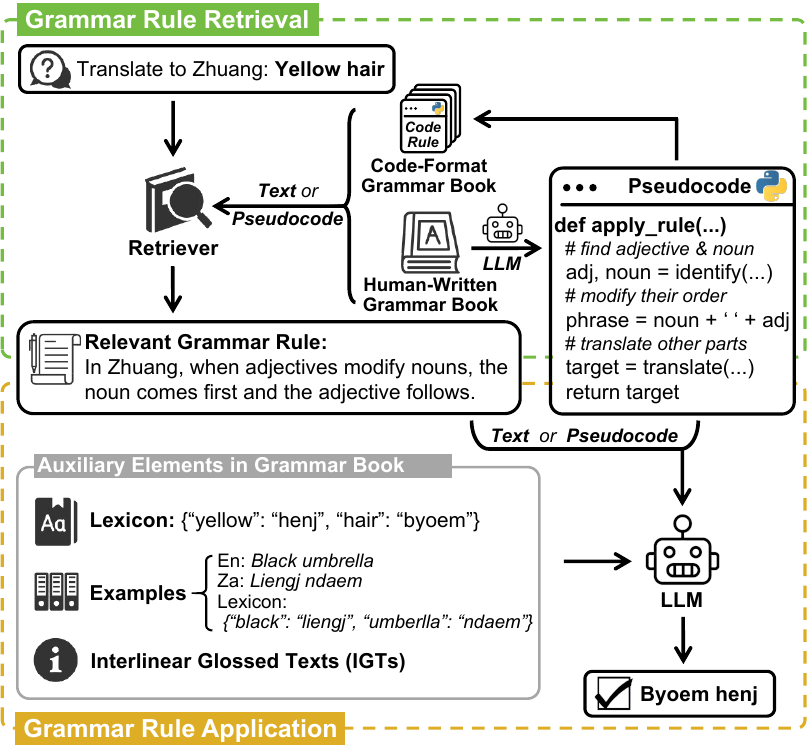}
\caption{An illustration of LLMs using grammar rules in code format to translate Zhuang, an extremely low-resource language.}
\label{fig:methodology}
\end{figure}

To address this gap, we introduce \textbf{\textsc{ZhuangRules}}, a fine-grained dataset focusing on the grammar rule understanding for Zhuang (ISO 639-1: za), a language frequently studied in XLR MT~\cite{zhang-etal-2024-teaching, bai2024longbench}. 
\textsc{ZhuangRules} decomposes grammar books into modular elements, consisting of 109 atomic grammar rules, each paired with an average of 5.6 Zhuang-Chinese parallel sentences for testing.
We ensure that correctly translating each test instance requires applying its corresponding grammar rule. 
We provide a Zhuang-Chinese lexicon for each sentence during testing, to disentangle grammar rule comprehension from lexical knowledge. 
These designs enable more controlled and interpretable evaluation in LLMs' ability of grammar understanding.

Given that each test sentence in \textsc{ZhuangRules} is annotated with its relevant grammar rule, we conduct a pilot study comparing two settings: providing all grammar rules (akin to an entire grammar book) versus supplying only the necessary one\. 
We find that the latter significantly outperforms the former, suggesting that grammar-based MT relies heavily on LLMs' ability to identify the required rules. 
We thus break down grammar-based MT into two stages, grammar rule retrieval and grammar rule application, and explore the following research questions:
(1) \textbf{RQ1:} Can LLMs retrieve the grammar rules required for translating a test sentence? (2) \textbf{RQ2:} Can LLMs effectively apply a given rule for translation as instructed?

We discover that grammar rule retrieval is a significant bottleneck in XLR MT, as LLMs struggle to effectively locate the necessary rules.  
Regarding rule application, we find LLMs can apply simple rules for translation when explicitly provided, with further gains observed when auxiliary elements like parallel sentence examples and interlinear glossed texts (IGTs) are included in the prompt. 
However, handling complex rules involving multiple actions remains a challenge for LLMs, with performance dropping to half that of simpler cases.

We further explore strategies to enhance LLMs’ ability to utilize grammar rules in these two steps. 
Inspired by prior works on improving LLM reasoning through code representations~\cite{liu-etal-2023-magic, li-etal-2024-eliciting-better}, we observe a strong analogy between the sequential operations in grammar rule application and the procedural structures of code. 
For instance, adding affixes to a word resembles an arithmetic addition operation, while selecting different affixes based on conditions aligns with an \texttt{if-else} structure in code. 
Therefore, as illustrated in Figure~\ref{fig:methodology}, we convert grammar rules into code-based representations using GPT-4o~\cite{hurst2024gpt}, to facilitate LLMs in translation. 
Additionally, we propose \textsc{Rule-by-Rule} retrieval, a simple but effective strategy that examines the necessity of each rule individually instead of processing the whole book directly.

Our experiments show that retrieving grammar rules in code format improves recall by 8.8\% compared to textual rules, and enhances LLMs' ability to utilize given rules effectively, boosting the translation performance by 12.2\% BLEU on \textsc{ZhuangRules}. 
This benefit is also observed on MTOB~\cite{tanzer2024a}, another translation benchmark for XLR MT. 
Finally, combining code rule with \textsc{Rule-by-Rule} retrieval strategy outperforms the end-to-end translation using the textual grammar book by 13.1\% BLEU on \textsc{ZhuangRules}.
 
Our contributions are summarized as follows:
(1) We underscore the necessity of breaking down grammar-based MT into two steps, rule retrieval and application, and identify rule retrieval as a major bottleneck. 
(2) We introduce a code-based format for grammar rules, improving LLMs' abilities in both steps and yielding substantial gains in translation performance.
(3) We present \textsc{ZhuangRules}, a dataset for explainable research on XLR MT using grammar rules, decomposing grammar books into structured elements including rules, parallel sentences, lexicons, and IGTs. 
Our data and code are publicly available to the community\footnote{\url{https://github.com/Infinite-set/ZhuangRules}}.

\section{Dataset: \textsc{ZhuangRules}}

We study the problem of grammar understanding using Zhuang, a low-resource language in China, which current LLMs hardly understand~\cite{zhang-etal-2024-teaching}.
We collect \textsc{ZhuangRules}, a set of 109 rules on Zhuang grammar written in Chinese.
Each rule is paired with several Zhuang-Chinese parallel phrases/sentences for testing, amounting to 608 pairs.
Each pair is further annotated with a bilingual lexicon covering all relevant lexical items in the sentences, which can disentangle the interference from LLMs' lack of Zhuang lexical knowledge when evaluating their understanding of grammar rules.
Compared to previous resources of complete grammar books~\cite{tanzer2024a,zhang-etal-2024-hire,hus-anastasopoulos-2024-back}, \textsc{ZhuangRules} enables more systematic and controllable analysis with its modularized structures for XLR MT.

\subsection{Rule Collection}
We collect the rules from two books on Zhuang written in Chinese, \zhsmall{《壮语通论》} (\textit{General Introduction to Zhuang Language}; \citealp{wei2006zhuangyutonglun}) and \zhsmall{《壮语基础教程》} 
(\textit{Basic Course of Zhuang Language}; \citealp{wei2008zhuangyujichujiaocheng}).
From these books, we collect grammar rules of Zhuang and their Zhuang-Chinese parallel phrases/sentences, which are typically concise and concretely illustrate the usage of the corresponding rule. 
For each sentence, we provide a Zhuang-Chinese lexicon covering the words appearing in the Zhuang sentence, which helps eliminate the interference to the experiments caused by the model not knowing word meanings.
The following is a grammar rule and one of its parallel examples. 
See data statistics and details of collection in Appendix~\ref{app:dataset}.

\advance\leftmargini -1em 
\begin{quote}
\small
    \textbf{Rule:} \zhsmall{在壮语中，形容词作名词的修饰时, 名词在前, 形容词于后。}(\textit{In Zhuang, when adjectives modify nouns, the noun comes first and the adjective follows.}) \\
    \textbf{Example:} \\
    \underline{Zhuang:} byoem henj \\
    \underline{Chinese:} \zhsmall{黄头发} (\textit{yellow hair}) \\
    \underline{Lexicon:} \{byoem: \zhsmall{头发}(\textit{hair}), henj: \zhsmall{黄}(\textit{yellow})\}
\end{quote}

\subsection{Analysis of Rules}

Zhuang exhibits diverse linguistic features in its grammar rules. To gain a deeper understanding of these rules, we annotate each with fine-grained attributes including \textbf{action}, \textbf{difficulty}, and \textbf{domain}. These attributes provide clear categorization, enabling detailed analysis of their utilization in XLR MT.

Regarding \textbf{action}, we identify the atomic operations required in applying each grammar rule for Chinese-Zhuang translation, such as adding affixes and reordering two words. 
The number of actions in a grammar rule can reflect its difficulty. 
We evaluate the \textbf{difficulty} of each rule based on the number of involved actions and the degree of difference between Zhuang and Chinese. 
The rules are thereby categorized into three levels: \textit{easy}, \textit{medium}, and \textit{hard}, with the average number of required operations being 1.2, 1.5 and 2.1, respectively. 
Additionally, we label rules according to their linguistic \textbf{domain}, following the taxonomy in WALS~\cite{wals}. 
We find that most rules in \textsc{ZhuangRules} deal with morphology and word order.
See Appendix~\ref{app:data_statistics} for detailed categorization of each attribute.

\begin{figure}[t]
\centering
\includegraphics[scale=0.36]{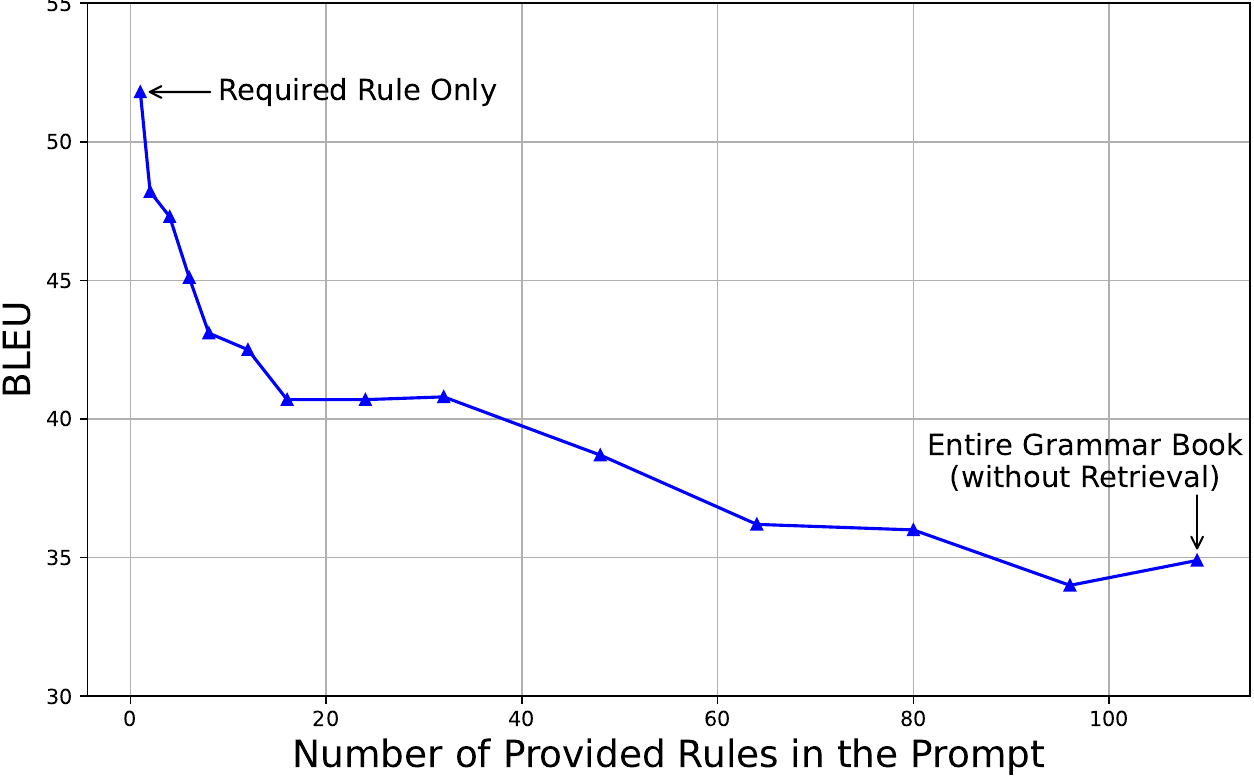}
\caption{Change in translation performance of providing varying numbers of irrelevant grammar rules in addition to the required rule (Qwen-2.5-72B-Instruct, Chinese $\rightarrow$ Zhuang).}
\label{fig:pilot_study}
\end{figure}

\section{Methodology}

Motivated by a pilot study that highlights the significant impact of the number of provided grammar rules on MT performance, we formalize the translation process using grammar books as a two-step procedure: grammar rule retrieval and grammar rule application.
To gain deeper insights into LLMs' capabilities, we evaluate their performance at each stage independently.
Additionally, we explore strategies to enhance LLMs' abilities in these two stages, especially representing grammar rules in a structured code-based format.

\subsection{Pilot Study: Providing LLMs with Varying Numbers of Rules}

Using \textsc{ZhuangRules}, we conduct a pilot study to assess whether providing varying numbers of grammar rules in the prompt affects LLMs' abilities of utilizing them for translation.

As shown in Figure~\ref{fig:pilot_study}, we observe that the translation performance declines sharply when we begin with only the required rule and progressively add more irrelevant rules\footnote{We observe similar trends in both Chinese-to-Zhuang and Zhuang-to-Chinese translation. See details in Appendix~\ref{app:varying_number_of_rules}.}. 
This suggests that LLMs struggle with numerous irrelevant rules in the grammar book, highlighting that the performance of grammar-based XLR MT is closely related to LLMs’ ability to identify the required grammar rules. 

Based on these findings, we are motivated to decouple rule retrieval from the process of end-to-end grammar-based translation, 
and evaluate two separate abilities of LLMs: finding the relevant grammar rules from a grammar book, and applying them in translation.

\subsection{Retrieving Grammar Rules}
Our pilot study indicates that the performance of grammar-based XLR MT is closely related to LLMs’ ability to identify the required grammar rules. 
Therefore, given a sentence to be translated, we first explore whether LLMs can find the required rules from a grammar book, which is simulated by the concatenation of all the rules in \textsc{ZhuangRules} (approximately 4K tokens).

We mainly explore whether changing the task format or presentation of grammar rules affects LLMs' abilities of finding required grammar rules.
As a baseline, we adopt \textbf{BM25}~\cite{robertson2009probabilistic}, where we use the test sentence as the query and retrieve the top $k$ relevant rules from the grammar book.

\paragraph{Changing the Task Format}
Similar to the default setting of performing XLR translation with a full book in previous works~\cite{tanzer2024a}, given a grammar book (a concatenation of rules) and a sentence to be translated as input, we instruct LLMs to output relevant rules in the book.
We refer to this strategy as \textbf{\textsc{Full-Book}}.

Considering the \textsc{Full-Book} approach places a high demand on the model's ability of long-context understanding, we propose another \textbf{\textsc{Rule-by-Rule}} strategy with a much shorter input length. 
In this approach, instead of handling the entire grammar book at once, LLMs examine whether each rule in the grammar book is relevant to the sentence to be translated individually.
Specifically, we input one candidate grammar rule and a test sentence at a time, and require LLMs to perform binary classification over their relevance.

\paragraph{Changing the Representation of Rules}
Previous works find that code formats can enhance the reasoning capabilities of LLMs in tasks involving math or logic~\cite{chen2023program,liu-etal-2023-magic}. 
These advantages stem from the similarities between these tasks and code representations in terms of the modular structure and control flow.
Similarly, operations involved in grammar rules exhibit a natural resemblance to code.
For example, adding or removing affixes resembles arithmetic operations, while selecting different affixes based on conditions can be represented with an \texttt{if-else} structure in code.
Building on this observation, we investigate whether the benefits of code-based reasoning extend to grammar rule understanding.

We convert the textual rules in \textsc{ZhuangRules} into \textbf{code rules} with GPT-4o~\cite{hurst2024gpt}.
Each code rule consists of two parts: (1) a concise comment outlining the steps for applying the rule in translation, and (2) a pseudo-code function that simulates the translation process. 
We only provide 5 exemplars for ICL during conversion, without applying further constraints on the structure or style of the pseudo-code since LLMs exhibit robustness to variations in code style and format when reasoning with code~\cite{liu-etal-2023-magic}. 
We randomly sample 10 code rules for quality check and find GPT-4o is effective at generating pseudocodes. 
All samples follow proper Python syntax, and only one sample omits minor information of the original rule. The remaining samples are all complete and accurate in both translation directions. 
See examples of code rules in Appendix~\ref{app:example_and_prompts}.

\subsection{Applying Grammar Rules}
After exploring the retrieval stage in XLR MT, we investigate LLM's ability to apply specific grammar rules in a targeted manner, by asking LLMs to translate a sentence directly using the required rule. 
We explore whether LLMs exhibit preferences for different representations of grammar rules and further examine the impact of auxiliary components commonly included in grammar books, such as parallel examples and IGTs.

\paragraph{Presentation of Rules}
We mainly investigate the effects of rule formats on grammar rule application, by comparing the settings using textual and code rules.
We additionally examine how the language used for writing rules affects translation, which is discussed in Appendix~\ref{app:varying_languages}.

\paragraph{Auxiliary Elements}
We further explore whether providing auxiliary elements from grammar books in the prompt can facilitate LLMs to apply rules.

\textbf{Parallel examples} can demonstrate the use of grammar rules more intuitively and are considered effective in XLR translation~\cite{court-elsner-2024-shortcomings,aycock2024can}.
In addition to the required rule, we provide each test sentence with several pairs of parallel sentences requiring the same rule.

\textbf{Interlinear glossed text} (IGT) is a line-by-line format for annotating linguistic corpora, where
each morpheme is labeled with a descriptive annotation.
It has been widely adopted as an intermediate form for low-resource language translation~\cite{ginn-etal-2024-teach,ginn-etal-2024-glosslm,ramos2024grammamt}.

As the two grammar books used for collecting \textsc{ZhuangRules} do not provide IGTs for the parallel sentences in them, we use GPT-4o to generate IGT annotations for each Zhuang sentence in \textsc{ZhuangRules}, with 123 IGTs collected from \zhsmall{《壮语语法标注文本》} (\textit{Annotated Zhuang Grammar Text}; \citealp{lan2016zhuangyuyufabiaozhu}) as ICL exemplars. 
We also conduct a quality check, finding that GPT-4o generates the correct symbols for 72\% of the morphemes.
See details in~\ref{app:zhuangrules_collection_details}.

In XLR translation experiments, we follow the approach described in~\citet{ramos2024grammamt}, where each parallel sentence in the prompt is paired with its corresponding IGT, and LLMs are instructed to first generate the IGT for the test sentence before translating.

\section{Experiments and Analyses}
We answer our two research questions by analyzing the current capabilities of LLMs and exploring the benefits of representing rules in a code-based format. 
We then aggregate our findings obtained from the two-step investigation to summarize the best practice for using grammar books in XLR MT.

\subsection{Experimental Setups}
\paragraph{Models}
We use three open-source LLMs for experiments: 
Qwen-2.5-7B-Instruct, Qwen-2.5-72B-Instruct~\cite{yang2024qwen2}, and Llama-3.1-70B-Instruct~\cite{dubey2024llama}.

\paragraph{Datasets}
\textsc{ZhuangRules} is the only dataset so far supporting controllable and interpretable experiments on grammar books. 
To validate the generalizability of our conclusions, we additionally use MTOB~\cite{tanzer2024a} for experiments, which contains a grammar book for Kalamang (ISO 639-3: kgv), an XLR language in Indonesia. 
We extract paragraphs of grammar descriptions and their corresponding examples from the book using regular expressions\footnote{The MTOB grammar book is written in a progressive style, with most example sentences requiring multiple rules for accurate translation. As we are not able to annotate all the required rules, we exclude MTOB from rule retrieval experiments and only use it for rule application experiments.}. See details of data construction from MTOB in Appendix~\ref{app:kalamang}.

\paragraph{Metrics}
For BM25, we report recall@$k$, reflecting whether the relevant rule appears in the top-$k$ retrieval results.
For \textsc{Full-Book} and \textsc{Rule-by-Rule} retrieval with LLMs, as the number of retrieved rules is not pre-defined, we report the average number and recall of retrieved rules.
For rule application, we report BLEU~\cite{papineni-etal-2002-bleu} and chrF++~\cite{popovic-2017-chrf}, using the implementation from \citet{post-2018-call}.

\paragraph{Prompting} 
For the experiments involving parallel sentence examples in grammar rule application, we randomly sample two pairs of examples for ICL from those requiring the same rule with the test instances.
For all settings except the ones explicitly annotated with \textbf{w/o Lexicon}, we include Zhuang-Chinese lexicons covering the words appearing in the prompt by default. 
See all the prompts used in our experiments in Appendix~\ref{app:example_and_prompts}.

\begin{table}[t]
\small
\centering
\setlength\tabcolsep{2.5pt}
\begin{tabular}{l|cc|cc}
\toprule
 & \multicolumn{2}{c|}{\textbf{za2zh}} & \multicolumn{2}{c}{\textbf{zh2za}}  \\
\midrule
\textbf{\textsc{Baseline}} & \textbf{rec@1} & \textbf{rec@5} & \textbf{rec@1} & \textbf{rec@5} \\
\midrule
BM25 & 26.3 & 41.6 & 13.5 & 27.3 \\
\midrule
\textbf{\textsc{Full-Book}} & \textbf{rec} & \textbf{\#rules} & \textbf{rec} & \textbf{\#rules} \\
\midrule
Qwen-2.5-7B & 3.1 & 2.3 & 5.2 & 2.2 \\
Llama-3.1-70B & 33.9 & 3.2 & 22.9 & 2.5 \\
Qwen-2.5-72B & 52.8 & 1.8 & 49.4 & 1.8 \\
\midrule
\textbf{\textsc{Rule-by-Rule}} & \textbf{rec} & \textbf{\#rules} & \textbf{rec} & \textbf{\#rules} \\
\midrule
Qwen-2.5-7B (text) & 55.1 & 2.5 & 67.9 & 4.0 \\
Qwen-2.5-7B (code) & 68.4 & 3.8 & 80.3 & 4.7 \\
Llama-3.1-70B (text) & 69.7 & 2.2 & 75.8 & 3.6 \\
Llama-3.1-70B (code) & 82.2 & 4.2 & \textbf{87.5} & 5.5 \\
Qwen-2.5-72B (text) & \underline{89.4} & 4.1 & 84.7 & 4.4 \\
Qwen-2.5-72B (code) & \textbf{89.6} & 3.9 & \underline{87.1} & 4.1 \\
\bottomrule
\end{tabular}
\caption{Performance of different rule retrieval strategies on \textsc{ZhuangRules}. The best scores are made \textbf{bold}, with the second \underline{underlined}.}
\label{tab:retrieval_one_book}
\end{table}

\begin{table*}[ht]
\small
\centering
\setlength\tabcolsep{3.5pt}
\begin{tabular}{l|ccc|ccc|c}
\toprule
\multirow{2}{*}{\textbf{BLEU / chrF++}} & \multicolumn{3}{c|}{\textbf{Zhuang} $\rightarrow$ \textbf{Chinese}} & \multicolumn{3}{c|}{\textbf{Chinese} $\rightarrow$ \textbf{Zhuang}} & \multirow{2}{*}{\textbf{Average}} \\
 & \textbf{Qwen (72B)} & \textbf{Llama (70B)} & \textbf{Qwen (7B)} & \textbf{Qwen (72B)} & \textbf{Llama (70B)} & \textbf{Qwen (7B)} &  \\
\midrule
\multicolumn{8}{c}{\textit{Baselines}} \\
\midrule
No Rule (w/o Lexicon)& 2.7 / 1.9 & 0.3 / 0.3 & 1.0 / 0.8 & 0.7 / 8.1 & 0.6 / 4.8 & 0.1 / 2.2 & 0.9 / 3.0 \\
No Rule  & 31.2 / 28.3 & 28.7 / 27.0 & 25.7 / 24.0 & 22.1 / 49.1 & 21.4 / 49.9 & 24.1 / 49.5 & 25.5 / 38.0 \\
Parallel Examples & 65.6 / 61.3 & 58.0 / 55.6 & 54.7 / 50.7 & 63.4 / 80.1 & 61.8 / 80.6 & 57.4 / 75.8 & 60.2 / 67.4 \\
\hspace{1em}+ Synthetic IGT & 65.6 / 62.6 & 60.4 / 60.3 & 56.3 / 55.7 & - / - & - / - & - / - & - / - \\ 
\midrule
\multicolumn{8}{c}{\textit{Textual Grammar Rules}} \\
\midrule
Random Textual Rule & 31.8 / 29.0 & 28.4 / 27.8 & 24.8 / 23.9 & 21.6 / 49.3 & 18.7 / 42.8 & 15.7 / 43.6 & 23.5 / 36.1 \\
Gold Textual Rule & 51.4 / 50.5 & 47.8 / 48.7 & 39.4 / 40.6 & 51.8 / 78.7 & 47.9 / 79.3 & 35.6 / 66.2 & 45.7 / 60.7 \\
\hspace{1em}+ Parallel Examples & \underline{70.7} / \underline{68.0} & \underline{68.4} / \underline{64.9} & 58.5 / 57.3 & \underline{80.7} / \underline{91.0} & \textbf{78.9} / \textbf{90.4} & \underline{63.9} / \underline{80.6} & \underline{70.2} / \underline{75.4} \\
\hspace{2em}+ Synthetic IGT & 67.9 / 66.6 & 62.2 / 64.0 & \underline{60.0} / 58.7 & - / - & - / - & - / - & - / - \\ 
\midrule
\multicolumn{8}{c}{\textit{Code Grammar Rules}} \\
\midrule
Random Code Rule & 27.7 / 25.9 & 23.7 / 25.2 & 22.3 / 22.8 & 20.2 / 48.7 & 18.4 / 48.0 & 15.2 / 44.0 & 21.3 / 35.8 \\
Gold Code Rule & 63.3 / 61.1 & 57.8 / 57.9 & 49.5 / 50.0 & 69.3 / 86.7 & 55.2 / 83.1 & 52.0 / 76.1 & 57.9 / 69.2 \\
\hspace{1em}+ Parallel Examples & \textbf{73.4} / \textbf{71.4} & \textbf{72.0} / \textbf{69.9} & \textbf{62.3} / \textbf{61.0} & \textbf{81.2} / \textbf{91.6} & \underline{77.8} / \underline{90.1} & \textbf{67.7} / \textbf{83.4} & \textbf{72.4} / \textbf{77.9} \\
\hspace{2em}+ Synthetic IGT & 68.5 / 66.8 & 65.1 / 64.4 & 59.8 / \underline{59.5} & - / - & - / - & - / - & - / - \\ 
\bottomrule
\end{tabular}
\caption{Translation performance of different settings of rule application on \textsc{ZhuangRules}. Note that IGTs do not support high-to-low-resource language translation, i.e. Chinese $\rightarrow$ Zhuang. The best scores are made \textbf{bold}, with the second \underline{underlined}.}
\label{tab:utilization}
\end{table*}

\begin{table}[t]
\small
\centering
\setlength\tabcolsep{2.5pt}
\begin{tabular}{l|cc|cc}
\toprule
& \multicolumn{2}{c|}{\textbf{kgv2eng}} & \multicolumn{2}{c}{\textbf{eng2kgv}} \\
& \textbf{BLEU} & \textbf{chrF++} & \textbf{BLEU} & \textbf{chrF++} \\ 
\midrule
No Rule (w/o Lex.) & 1.6 & 9.3 & 0.8 & 8.6 \\
No Rule & 12.0 & 34.6 & 39.9 & 63.8 \\
\midrule
Random Textual Rule & 11.9 & 34.8 & 40.7 & 63.5 \\ 
Random Code Rule & 13.3 & 36.3 & 37.9 & 61.9 \\
Gold Textual Rule & 14.6 & 39.2 & 43.8 & \textbf{67.3} \\ 
Gold Code Rule & \textbf{16.0} & \textbf{40.7} & \textbf{44.5} & 67.0  \\
\bottomrule
\end{tabular}
\caption{Translation performance of Qwen-2.5-72B-Instruct between English and Kalamang, using different rule formats.}
\label{tab:kalamang_utilization}
\end{table}

\subsection{RQ1: Can LLMs Find the Required Grammar Rule?}

\vspace{1mm}
\noindent\textbf{LLMs struggle to retrieve rules from \textit{the full grammar book} directly.}
In Table~\ref{tab:retrieval_one_book}, we compare the performance of different strategies for finding relevant rules for the sentence to be translated. 
BM25, relying solely on lexical overlap, fails to retrieve the relevant rules for more than half of the test instances within the top-5 results, underscoring the complexity of rule retrieval.

Using LLMs to identify the required rules from the entire grammar book (\textsc{Full-Book}) shows notable improvements over BM25, as LLMs can better capture the semantic relationships between rules and testing instances beyond simple lexical matching. 
However, their performance is still far from perfect, with the best results still hovering around 50\% in recall. 
Moreover, this approach is highly dependent on model capabilities, as weaker models, such as Qwen-2.5-7B-Instruct, are almost unable to find the correct rules.

This finding indicates that when LLMs perform end-to-end translation with a grammar book, they mostly do not know which grammar rules are necessary for translating a given sentence, in line with our pilot study in Figure~\ref{fig:pilot_study}, where we find a strong correlation between rule retrieval performance and final translation quality. 
By addressing the bottleneck of rule retrieval, it is possible to further improve the translation performance of LLMs.
We then explore two strategies for improving rule retrieval: transforming the task format and changing the form of rules.

\vspace{1mm}
\noindent\textbf{Converting the retrieval task into \textsc{Rule-by-Rule} classification helps. }
As shown in Table~\ref{tab:retrieval_one_book}, instead of providing the entire grammar book (\textsc{Full-Book}), examining each rule individually (\textsc{Rule-by-Rule}) leads to significantly better performance. 
This approach achieves nearly 80\% recall with fewer than 5 retrieved rules on average, making it a more practical solution compared to providing LLMs with the entire book (\textsc{Full-Book}).

\vspace{1mm}
\noindent\textbf{Rules in code format are more LLM-friendly for retrieval.}
As shown in Table~\ref{tab:retrieval_one_book}, converting textual rules into code forms, combined with the \textsc{Rule-by-Rule} strategy, improves retrieval performance across all models, achieving up to 90\% recall while maintaining a manageable number of retrieved rules.
Code forms transform descriptive rules into procedural knowledge, making it easier for LLMs to understand the requirements of translation than textual rules.

\subsection{RQ2: Can LLMs Apply a Given Grammar Rule as Instructed?}

\vspace{1mm}
\noindent\textbf{LLMs can apply simple rules for translation when explicitly given. }
As shown in Table~\ref{tab:utilization}, providing the necessary textual grammar rule significantly improves translation performance over providing no or random grammar rules.
For example, on \textsc{ZhuangRules} we observe an absolute increase of 26\% chrF++ in Qwen-2.5-72B's performance, after providing the gold grammar rules. 
This finding is also validated in another XLR language, Kalamang. 
As shown in Table~\ref{tab:kalamang_utilization}, there is an increase of 4\% chrF++ on Kalamang-English translation after providing relevant rules, compared to the settings providing no or random rules.
These improvements can be attributed to LLMs’ strong capability to comprehend and follow clearly-given instructions.

We then investigate the role of other auxiliary elements of the grammar books in facilitating the understanding of grammar rules. 
In the absence of explicit grammar rules, parallel examples and IGTs prove valuable for translation, aligning with findings from previous work~\cite{aycock2024can,ginn-etal-2024-teach}.
After pairing textual grammar rules with parallel examples, we observe an average chrF++ gain of 14.7\% over using rules alone. 
We conjecture that parallel examples help LLMs identify common patterns or usages in a more intuitive way than rules alone. 
Besides, their format resembles testing instances, aiding LLMs in better mimicking the translation process.
However, further incorporating synthetic IGTs when grammar rules are provided reduces the gain from parallel examples.
This may be due to the noise introduced by the GPT-4o generated IGTs (see Appendix~\ref{app:zhuangrules_collection_details}), which may be inconsistent with the provided rules.

\vspace{1mm}
\noindent\textbf{Code formats enhance rule understanding, especially for difficult rules.}
As shown in Table~\ref{tab:utilization}, code rules consistently outperform textual rules, resulting in an average improvement of 8.5\% chrF++ on \textsc{ZhuangRules}.
This trend holds for Kalamang as well, as seen in Table~\ref{tab:kalamang_utilization}, where code rules also outperform textual rules in most settings. 

The advantage of code format is more pronounced on difficult grammar rules, which involve more operation steps when applied for translation.
As shown in Table~\ref{tab:difficulty}, the improvement of code rules over textual ones is particularly noticeable on the \textit{hard} subset of \textsc{ZhuangRules}, especially when parallel examples are unavailable. 
See Appendix~\ref{app:case_study} for a case study.

Beyond performance improvements, code formats exhibit more advantages as a structural representation for XLR translation.
It offers a language-agnostic interface for understanding grammar, without requiring pre-defined language-specific protocols such as the glossing conventions required by IGTs~\cite{rules2008conventions}.
Furthermore, the conversion into code format can be well performed by powerful LLMs, requiring little human labor while yielding substantial performance gains.

\begin{table}[t]
\small
\centering
\setlength\tabcolsep{1.5pt}
\begin{tabular}{l|rcr|rcr}
\toprule
& \multicolumn{3}{c|}{\textbf{za2zh}} & \multicolumn{3}{c}{\textbf{zh2za}} \\
& \textit{easy} & \textit{medium} & \textit{hard} & \textit{easy} & \textit{medium} & \textit{hard} \\
\midrule
No Rule & 29.5 & 28.8 & 24.2 & 18.6 & 22.2 & 30.5 \\
\midrule
Text Rule & 65.6 & 51.3 & 34.6 & 85.5 & 82.4 & 69.3 \\
Code Rule & 76.3 & 57.9 & 48.6 & 93.0 & 87.5 & 76.8 \\
$\Delta$ & +10.7 & +6.6 & \textbf{+14.0} & \textbf{+7.5} & +5.2 & \textbf{+7.5} \\
\midrule
Text Rule + 2 Ex. & 78.8 & 69.1 &  52.0 &  95.2 & 91.1 & 85.6 \\
Code Rule + 2 Ex. & 81.1 & 72.6 &  56.7 &  94.9 &  92.0 &  87.2 \\
$\Delta$ & +2.3 & +3.5 & \textbf{+4.6} & +0.3 & +1.0 & \textbf{+1.6} \\
\bottomrule
\end{tabular}
\caption{Translation performance (chrF++) of Qwen-2.5-72B-Instruct on \textsc{ZhuangRules} in different levels of difficulty. We additionally report the performance difference ($\Delta$) between using textual and code rules. }
\label{tab:difficulty}
\end{table}

\begin{figure}[t]
\centering
\includegraphics[width=0.9\columnwidth]{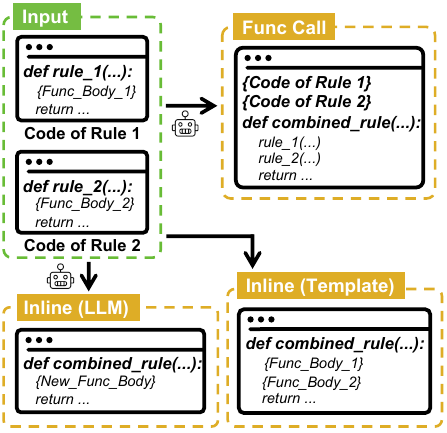}
\caption{An illustration of three strategies for combining multiple rules.}
\label{fig:multi_code}
\end{figure}

\begin{table}[t]
\small
\centering
\setlength\tabcolsep{2.5pt}
\begin{tabular}{l|cc|cc}
\toprule
& \multicolumn{2}{c|}{\textbf{za2zh}} & \multicolumn{2}{c}{\textbf{zh2za}} \\
& \textbf{BLEU} & \textbf{chrF++} & \textbf{BLEU} & \textbf{chrF++} \\ 
\midrule
No Rule & 30.2 & 25.2 & 20.9 & 47.9 \\
One Textual Rule & 39.5 & 31.6 & 39.5 & 64.0 \\
Two Textual Rules  & 47.3 & 41.7 & 45.5 & 73.9 \\
\midrule
\textsc{Func Call} & 46.6 & 44.7 & 50.6 & 72.3 \\
\textsc{Inline} (Template) & 56.0 & 48.5 & \textbf{51.4} & \textbf{75.5} \\
\textsc{Inline} (LLM) & \textbf{56.7} & \textbf{52.7} & 50.4 & 74.7 \\
\bottomrule
\end{tabular}
\caption{Translation performance of Qwen-2.5-72B using different strategies for utilizing multiple grammar rules.}
\label{tab:multi_rules}
\end{table}

\vspace{1mm}
\noindent\textbf{Code formats support simultaneous application of multiple rules.}
In the previous experiments, we adopt an idealized setting where translating a testing sentence requires only a single grammar rule, to facilitate controlled experiments. 
However, in real-world scenarios, translating a sentence often necessitates the application of multiple grammar rules. 
To validate our findings under a more realistic setting, we collect a small set of 96 testing instances requiring multiple rules.
Concretely, we select suitable candidates from the parallel sentences in \textsc{ZhuangRules} and modify them to require two rules for translation. See Appendix~\ref{app:dataset} for details.

Using code formats for multiple grammar rules in translation is non-trivial. 
The structural nature of code enables diverse strategies for combining these rules.
As shown in Figure~\ref{fig:multi_code}, we explore three combining strategies. 
In the \textsc{Func Call} strategy, LLMs generate a new function that determines the order of rule application and calls two existing rule functions sequentially.
For the other two \textsc{Inline} strategies, the bodies of two rules compose the new function. In \textsc{Inline} (Template), we directly concatenate the bodies of two rules in random order into a new function, while in \textsc{Inline} (LLM), LLMs freely generate a new function combining the contents of two rule functions.

As shown in Table~\ref{tab:multi_rules}, providing all necessary rules significantly improves translation performance compared to providing no or one rule, demonstrating that LLMs can effectively apply multiple rules.
Consistent with the findings in single-rule application, code rules yield noticeable advantages over textual ones.
Among the strategies for combining multiple code rules, \textsc{Inline} (LLM) achieves the best overall performance, indicating that LLMs can find a more effective way of organizing multiple rules by themselves.

\subsection{Putting it Together: Best Practice of Utilizing Grammar Rules}
In previous experiments, we decompose the grammar-based translation process into two steps and examine them independently. 
We find that although LLMs show promising abilities to apply it for translation when given a relevant grammar rule, they struggle to find relevant rules from a long grammar book for a sentence to be translated. 
To mitigate the bottleneck of rule retrieval and enhance the application of harder rules, we propose a \textsc{Rule-by-Rule} retrieval strategy and represent rules in code format. These approaches enhance the performance of both steps individually.

Now we present a comprehensive evaluation of the entire pipeline, assessing various combinations of design choices for each step. 
As shown in Table~\ref{tab:together}, the gains in rule retrieval brought by \textsc{Rule-by-Rule} retrieval strategy and code-based rule representations can propagate to the final translation performance, with up to 13.1\% increase in chrF++ for Qwen-2.5-72B-Instruct compared to feeding the whole grammar books to LLMs, the default practice of previous works.
Based on these findings, we summarize a promising strategy of using grammar books with a moderate number of grammar rules: \textbf{first, converting the rules in the grammar book into code format, then employing \textsc{Rule-by-Rule} retrieval with code rules, and finally applying the retrieved code rules for translation}.

\begin{table}[t]
\small
\centering
\setlength\tabcolsep{3.6pt}
\begin{tabular}{l|cc|cc}
\toprule
\multirow{2}{*}{\diagbox{\textbf{Retrieval}}{\textbf{Application}}} & \multicolumn{2}{c|}{\textbf{Text Rule}} & \multicolumn{2}{c}{\textbf{Code Rule}} \\
& \textbf{za2zh} & \textbf{zh2za} & \textbf{za2zh} & \textbf{zh2za} \\
\midrule
w/o Retrieval & 42.6 & 64.2 & - & - \\
\textsc{Full-Book} & 43.1 & 66.5 & 50.2 & 69.4 \\
\textsc{Rule-by-Rule} (text) & 46.8 & 69.9 & 55.4 & 72.3 \\
\textsc{Rule-by-Rule} (code) & 47.6 & 71.4 & \textbf{55.7} & \textbf{74.3} \\
\bottomrule
\end{tabular}
\caption{Translation performance (chrF++) of Qwen-2.5-72B-Instruct in different settings of retrieving and applying rules. The result of \textit{w/o Retrieval + Code Rule} is not reported because the input exceeds the maximum context length of evaluated LLMs.}
\label{tab:together}
\end{table}

\section{Related Works}
\paragraph{XLR MT with LLMs}
LLMs have been widely adopted for XLR MT for their training-free nature and efficient use of limited data.
Existing works focusing on improving XLR MT performance using linguistic resources, such as dictionaries~\cite{ghazvininejad2023dictionarybasedphraselevelpromptinglarge,elsner-needle-2023-translating,lu-etal-2024-chain,dimakis-etal-2024-dictionary}, parallel corpra~\cite{court-elsner-2024-shortcomings,guo-etal-2024-teaching,liu2024an,zhang-etal-2024-teaching}, grammar rules~\cite{tanzer2024a,team2024gemini,zhang-etal-2024-hire,hus-anastasopoulos-2024-back,aycock2024can}, and IGTs~\cite{ginn-etal-2024-teach,ginn-etal-2024-glosslm,ramos2024grammamt}.
However, these works arrive at divergent conclusions, due to inconsistent experimental setups. 
In contrast, we conduct controlled experiments using decomposable data, unveiling LLMs' true capabilities in XLR MT.

\paragraph{Reasoning with Code}
Code has been proven a better interface for LLMs to perform reasoning than natural languages under many circumstances such as math reasoning~\cite{li-etal-2024-eliciting-better} and causal reasoning~\cite{liu-etal-2023-magic}. 
Existing works either use the modularity feature of code to describe structures~\cite{madaan-etal-2022-language,bogin-etal-2024-leveraging}
or use the control flow of code to describe logical flows~\cite{liu2025eliciting, puerto-etal-2024-code,chae-etal-2024-language}. 
We take advantage of both aspects of code to represent grammar rules, which greatly facilitate LLMs in XLR MT.

In the area of machine translation, rule-based machine translation systems, such as Apertium~\cite{Forcada2011ApertiumAF}, employ structured rule specification languages. 
Our work on using code for machine translation extends this idea, adapting the concept of structured rules within the context of LLMs.

\section{Conclusion}
We present a comprehensive evaluation of LLMs' ability to leverage grammar books for translating XLR languages. Using \textsc{ZhuangRules}, a newly introduced dataset for interpretable research, we decompose the translation process into two key steps: grammar rule retrieval and grammar rule application. Our analysis reveals that while LLMs can effectively apply simple grammar rules, they struggle significantly with retrieving the relevant rules. To address this, we propose to represent grammar rules in code format, which improves the performance of both retrieval and application, eventually enhancing the overall translation quality.

\section*{Limitations}
\paragraph{Scope of Studied Languages} 
Collecting suitable data for controlled experiments on extremely low-resource languages requires significant effort. We do our best to investigate the research question on two XLR languages, Zhuang and Kalamang, which are considered unseen by current LLMs and thus frequently used for the research of on-the-fly XLR MT with linguistic resources~\cite{tanzer2024a,aycock2024can,zhang-etal-2024-teaching,bai2024longbench}.
Moreover, the substantial typological differences between Zhuang (morphologically analytic, SVO) and Kalamang (morphologically synthetic, SOV) lend credence to the generalizability of our findings.

Unlike other low-resource languages that LLMs have some preliminary support for, with a sufficiently large corpus available for fine-tuning, we believe that these extremely low-resource languages could benefit more from grammar book-based approaches. 
Thus, we select Zhuang and Kalamang as our primary research targets. We encourage future work to explore whether our findings generalize to low-resource languages with slightly more data availability.

\paragraph{Token Efficiency}
While code rules substantially enhance LLMs' understanding of grammar, they are longer than textual rules, resulting in a higher token count in the input. 
However, this issue can be mitigated through effective rule retrieval over the grammar book. As demonstrated in our experiments, the best retrieval strategy not only improves translation performance but also maintains a manageable input length.

\paragraph{Idealized Scenarios}
In most of our experiments, test instances are designed to require only a single grammar rule for translation. This controlled setting allows for a more precise and interpretable diagnosis of LLMs' behavior. 
However, this represents an idealized scenario, as real-world translations often require applying multiple rules from grammar books. 
To bridge this gap, we have conducted preliminary experiments on a subset of instances requiring two rules, further validating our findings. 
We leave the exploration of more open-ended and unconstrained translation settings for future work.

\section*{Acknowledgements}
This work is supported in part by NSFC~(62161160339) and Beijing Science and Technology Program~(Z231100007423011). We thank the anonymous reviewers for their valuable feedback. 
We also thank Shaodan Sui for the efforts in the dataset construction.
For any correspondence, please contact Yansong Feng.

\bibliography{anthology,custom}

\clearpage
\appendix

\section{Introduction of Studied Languages}
\label{app:languages}

\paragraph{The Zhuang Language} 
Zhuang is a group of Kra–Dai languages spoken primarily by the Zhuang people in Southern China, particularly in Guangxi Province and the neighboring regions of Yunnan and Guangdong. With over 16 million speakers, Zhuang is one of the largest minority languages in China. 
Zhuang is an isolating language, characterized by a lack of inflectional morphology.
The current official writing system for Zhuang is the Latin script.  
In this study, we focus on Standard Zhuang, the officially standardized form of the language. Notably, current open-source and commercial language models show near-zero proficiency in processing Zhuang, highlighting its status as an underrepresented language that lacks support in existing LLMs.

\paragraph{The Kalamang Language} 
Kalamang is an endangered language primarily used in the villages on the largest of the Karas islands off the west coast of the Bomberai Peninsula in Indonesian Papua. 
It belongs to the Greater West Bomberai language family. 
Kalamang has a small speaker population (less than 200 people) and was only recently documented~\cite{da72812684ef4a0c9d1476e6357c18c2, tanzer2024a}. 
Written Kalamang uses the Latin script.

\section{Dataset Collection and Statistics}
\label{app:dataset}

\subsection{Grammar Book Extracts}
\begin{figure*}[t]
\centering
\includegraphics[scale=1.0]{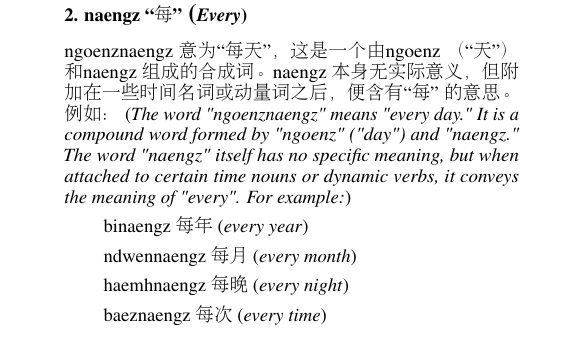}
\caption{A simplified extract from the Zhuang grammar book. The \textit{text in italics} are the English translations.}
\label{fig:app:za_extract}
\end{figure*}

\begin{figure*}[t]
\centering
\includegraphics[scale=0.85]{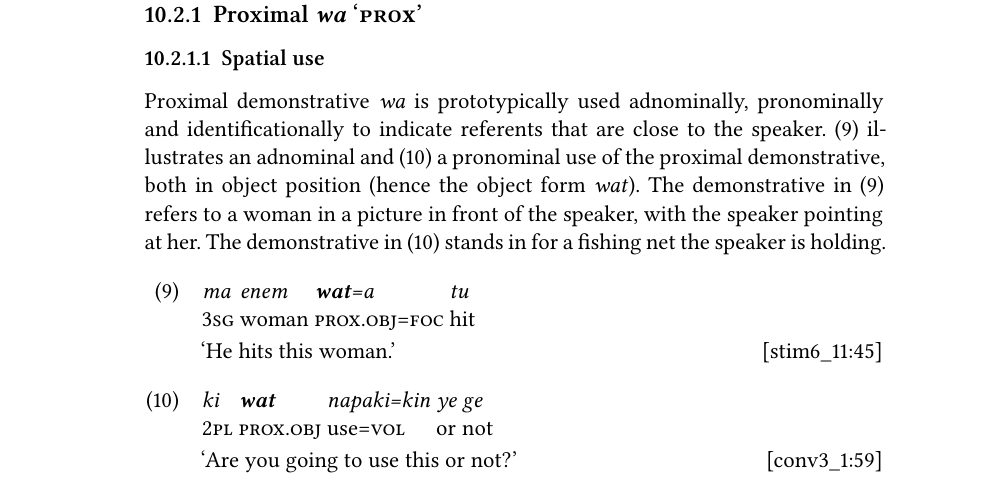}
\caption{A simplified extract from the Kalamang grammar book.}
\label{fig:app:kgv_extract}
\end{figure*}

In Figure~\ref{fig:app:za_extract}, we provide a brief extract from the Zhuang grammar book~\cite{wei2006zhuangyutonglun,wei2008zhuangyujichujiaocheng}, and in Figure~\ref{fig:app:kgv_extract} an extract from the Kalamang grammar book~\cite{da72812684ef4a0c9d1476e6357c18c2}. Each figure begins with a textual grammar rule, followed by a set of parallel phrases or sentences as examples.

\subsection{Collection Details of Zhuang Data}
\label{app:zhuangrules_collection_details}

\paragraph{Rule Selection of \textsc{ZhuangRules}} 
From two grammar books, we extract suitable grammatical rules and their parallel sentence examples.
A suitable rule should be concise and relatively independent, which means it does not rely on other rules as prerequisites.
Afterward, we consult the dictionary at the end of the grammar book to build a lexicon for each Zhuang sentence in the dataset. Finally, we annotate fine-grained attributes for each rule, including action, difficulty, and domain, which will be discussed in the dataset statistics (Appendix~\ref{app:data_statistics}).

\paragraph{IGT Generation for \textsc{ZhuangRules}} 
We use GPT-4o to generate an IGT for each Zhuang sentence in \textsc{ZhuangRules}.
To obtain ICL examples, we collect 123 pieces of IGT from \zhsmall{《壮语语法标注文本》} (\textit{Annotated Zhuang Grammar Text}; \citealp{lan2016zhuangyuyufabiaozhu}). 
Each data point consists of a Zhuang sentence, its IGT, and its Chinese translation.
The following is an example.

\begin{small}
\begin{exe}
  \ex
  \gll gou  aeu  aen  laj \\
       1\textsc{SG} \zhsmall{要}(\textit{want})  \textsc{CL}-\zhsmall{个}(\textit{one})  \zhsmall{下面}(\textit{below}) \\
  \glt ``\zhsmall{我要下面那个} (\textit{I want the one below})''  
\end{exe}
\end{small}


We then use this collection of 123 IGT examples, together with a gloss list of 67 symbols, to guide GPT-4o in annotating each sentence in \textsc{ZhuangRules}.

We sample 20 IGTs generated by GPT-4o and check their quality. 
GPT-4o generates the correct symbols for 72\% of the morphemes.

\begin{figure*}[t]
\centering
\includegraphics[scale=0.90]{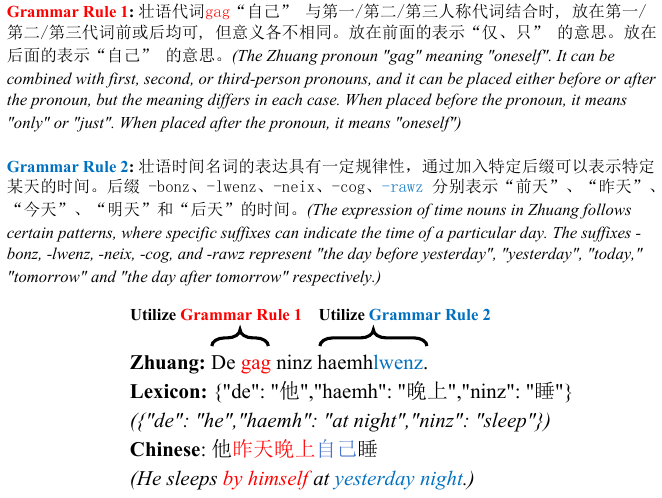}
\caption{An testing instance that requires multiple grammar rules. To construct this, we modify sentences governed by Rule 1 \textit{(a rule about word order)} by inserting words related to Rule 2 \textit{(a morphological rule)}. The \textit{text in italics} are the English translations.}
\label{fig:app:multirules_example}
\end{figure*}

\paragraph{Construction of Multi-Rule Testing Instances}
We additionally collect a small set of data (96 test instances in total) from \textsc{ZhuangRules} that requires multiple rules for experiments. Specifically, we select suitable candidates from the parallel sentences in \textsc{ZhuangRules} and modify them to require two rules for translation. 
For instance, we modify sentences governed by a word order rule by inserting or replacing words related to a morphological rule. See Figure~\ref{fig:app:multirules_example} for an example.

\paragraph{Data Checking}
We check the data in \textsc{ZhuangRules} to ensure it contains no information that names or uniquely identifies individuals, nor any offensive content.

\subsection{Collection Details of Kalamang Data} 
\label{app:kalamang}
\paragraph{Collection Process}
The Kalamang grammar book in MTOB is written in a progressive style, where the understanding of most example sentences relies on multiple rules, which are difficult to annotate.

Given the infeasibility of engaging proficient
Kalamang speakers in annotation, we leverage regular expressions to automatically match and extract certain contents from the book. 
Specifically, most parallel sentence examples in MTOB follow a three-line structure: the original Kalamang sentence, its IGT, and its English translation. We write regular expressions to capture this structure, using the paragraph preceding it as the corresponding grammar rule.

However, because the MTOB grammar book often requires multiple rules for each example sentence, and lacks annotations for specific rules, we exclude it from rule retrieval experiments. Additionally, the small number of parallel sentence examples (1.6 per rule on average) limits its utility in rule application experiments. Consequently, we focus primarily on grammar rules due to inadequate in-context learning sentence examples.

\paragraph{Potential Issues}
Note that the collected data might contain noise from two sources. 
First, many rules and their parallel examples in MTOB are not well-structured and cannot be captured by regular expressions.
Second, rules in the Kalamang grammar book are often written in a progressive style (i.e., they directly reference parts of previously defined grammar rules without providing additional clarification, making it difficult to understand their meaning when viewed in isolation without giving those previously defined rules) and include extensive explanations for some parallel sentence examples, which complicates the rule descriptions.

These issues might hinder the performance of using collected grammar rules for translation, explaining why including grammar rules leads to a smaller improvement in Kalamang than in Zhuang.

\subsection{Data Statistics}
\label{app:data_statistics}

\paragraph{Sizes and Lengths}
\textsc{ZhuangRules} contains 109 rules in total, with an average length of 57.1 Chinese characters. 
These rules are paired with 608 Zhuang-Chinese parallel examples, of which 432 are words or phrases, and 176 are sentences. 
The average length of sentence examples is 8.4 Chinese characters or 5.4 Zhuang words. 
We also calculated the total token number for each element in \textsc{ZhuangRules} using the tokenizer for Qwen-2.5-72B-Instruct. The results are shown in Table~\ref{tab:app:token_statistics}.

From MTOB, we collect 97 rules for Kalamang, with an average length of 120.4 English words. 
These rules are paired with 152 Kalamang-English parallel examples. The average length of sentence examples is 9.2 English words or 10.9 Kalamang words.

\paragraph{Additional Attributes}
For each rule in \textsc{ZhuangRules}, we annotate several fine-grained attributes to gain deeper insights, including action, difficulty, and domain.

\begin{itemize}
    \item \textbf{Action:} For each rule, we annotate the actions that are required when applying this rule for Chinese-Zhuang translation. Possible actions include \textit{add} (adding affixes to form a new word or adding some words to form a new phrase), \textit{delete} (omitting certain words), \textit{reorder} (reordering several words), \textit{break} (breaking a word to several parts), and \textit{select} (selecting a branch in the rule). As shown in Table~\ref{tab:app:data_statistics}, the two most common actions are \textit{add} and \textit{reorder}, which nearly half of the rules require. 
    \item \textbf{Difficulty:} We categorize the rules into three levels: \textit{easy}, \textit{medium} and \textit{hard}, based on the number of actions it involves and the degree of difference between Zhuang and Chinese reflected by this rule. Out of the 109 rules, there are 47 \textit{easy}, 43 \textit{medium}, and 19 \textit{hard} rules. The average number of required operations is 1.2, 1.5, and 2.1 for the three levels of rules, respectively. 
    \item \textbf{Domain:} Following WALS~\cite{wals}, a database of structural properties of languages, we label the rules with its linguistic domain, including \textit{morphology}, \textit{nominal categories}, \textit{nominal syntax}, \textit{verbal categories}, \textit{word order}, \textit{simple clauses}, \textit{complex sentences}, and \textit{lexicon}. As shown in Table~\ref{tab:app:data_statistics}, most of the rules deal with how the words are formed and ordered.
\end{itemize}

\begin{table}[t]
\small
\centering
\setlength\tabcolsep{7pt}
\begin{tabular}{lr|lr}
\toprule
\textbf{Action} &   & \textbf{Domain} & \\
\midrule
Add & 53 & Morphology & 28 \\
Delete & 6 & Nominal Categories & 16\\
Reorder & 54 & Nominal Syntax & 1 \\
Break & 6 &  Verbal Categories & 14 \\
Select & 22 & Word Order & 37 \\
& & Simple Clauses & 7  \\
& & Complex Sentences & 1 \\
& & Lexicon & 6 \\
\bottomrule
\end{tabular}
\caption{Number of rules for each type of action or domain in \textsc{ZhuangRules}.}
\label{tab:app:data_statistics}
\end{table}

\begin{table}[t]
\small
\centering
\setlength\tabcolsep{1.5pt}
\begin{tabular}{l|r}
\toprule
\textbf{Element} & \textbf{Total Tokens} \\
\midrule
Textual Rules & 4386 \\
Code Rules \textit{(Ver. For Retrieval)} & 23,165 \\
Code Rules  & 32,231 \\
\midrule
Lexicon & 18819 \\
Parallel Examples & 6,992 \\
IGTs & 4,927 \\
\midrule
Text Rules + Lex. + 2 Ex.  & 13,973 \\
Text Rules + Lex. + 2 Ex. + IGT & 15,787 \\
Code Rules + Lex. + 2 Ex. & 41,818 \\
Code Rules + Lex. + 2 Ex. + IGT & 43,632 \\
\bottomrule
\end{tabular}
\caption{Number of tokens for each elements in \textsc{ZhuangRules}.}
\label{tab:app:token_statistics}
\end{table}

\section{Implementation Details}
\label{app:example_and_prompts}

\paragraph{Conversion of Code Rules}
Given our decomposition of using grammar books into two distinct steps—rule retrieval and rule application, we design different styles of code for the two steps, following the principle of using task-appropriate data. 
The code for rule retrieval shows the steps to examine whether the rule should be applied, while the code for rule application demonstrates the translation process. 
We provide an example of the code rules for rule retrieval in Table~\ref{tab:app:code_grammar_rule_retrieval}, and for rule application in Table~\ref{tab:app:code_grammar_rule_utilization}.

\paragraph{Prompts for Rule Retrieval and Application} 
For \textsc{Rule-by-Rule} retrieval, we provide an example of prompt in Table~\ref{tab:app:prompt_retrieval}. 
For rule application, we provide a prompt example in Table~\ref{tab:app:prompt_utilization}. 
We typically employ zero-shot learning without chain-of-thought~\cite{wei2023chainofthoughtpromptingelicitsreasoning} for efficiency considerations.

\paragraph{Hyperparameters of Prompting} 
For the prompt-based method, we use greedy search without doing a hyperparameter search.

\section{Additional Experimental Results}

\subsection{Effect of Providing Varying Numbers of Rules}
\label{app:varying_number_of_rules}
\begin{figure*}[t]
\centering
\includegraphics[scale=0.47]{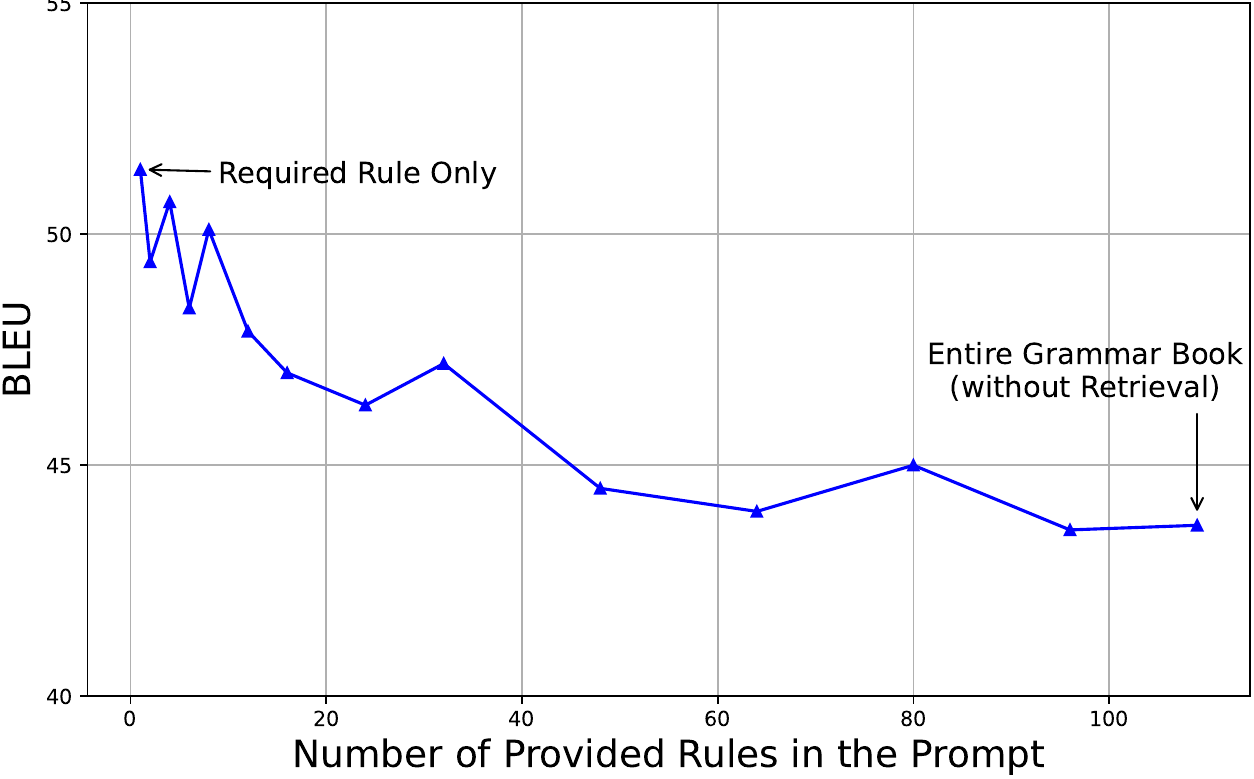}
\caption{Performance of BLEU by rule number (Qwen-2.5-72B-Instruct, Zhuang $\rightarrow$ Chinese)}
\label{fig:app:correlation_za2zhbleu}
\end{figure*}

We present the complete results of MT performance by providing varying numbers of rules for Qwen-2.5-72B-Instruct. 
For Chinese-to-Zhuang translation, the BLEU score is shown in Figure~\ref{fig:pilot_study}.
For Zhuang-to-Chinese translation, the BLEU score is shown in Figure~\ref{fig:app:correlation_za2zhbleu},
We observe similar phenomena on both translation directions.

\begin{table}[t]
\small
\centering
\setlength\tabcolsep{2.5pt}
\begin{tabular}{l|cc|cc}
\toprule
& \multicolumn{2}{c|}{\textbf{za2zh}} & \multicolumn{2}{c}{\textbf{zh2za}} \\
& \textbf{BLEU} & \textbf{chrF++} & \textbf{BLEU} & \textbf{chrF++} \\ 
\midrule
No Rule & 28.2 & 24.4 & 21.7 & 51.4 \\
\midrule
\multicolumn{5}{l}{\textit{Grammar Rules Only}} \\
\midrule
Chinese Rule & 51.4 & 50.5 & 51.8 & 78.7 \\ 
English Rule & 51.4 & 48.9 & 48.4 & 78.3 \\
\midrule
\multicolumn{5}{l}{\textit{Grammar Rules with Parallel Examples}} \\
\midrule
Chinese Rule + 2 Ex. & \textbf{70.7} & \textbf{68.0} & 80.7 & 91.0 \\
English Rule + 2 Ex. & 66.2 & 62.3 & \textbf{81.6} & \textbf{92.3} \\
\bottomrule
\end{tabular}
\caption{MT performance of Qwen-2.5-72B-Instruct on Zhuang grammar rules written in different languages.}
\label{tab:app:english_rules}
\end{table}

\subsection{Effect of Grammar Rule Languages}
\label{app:varying_languages}
As previous work shows that LLM performance varies when instructions are presented in different languages~\cite{shi2023language,etxaniz-etal-2024-multilingual}, we examine how the language of rules affects translation.
We translate the Chinese rules in \textsc{ZhuangRules} into English with GPT-4o and revise the translation manually. 

We present the results of using the grammar rules written in English for translation in Table~\ref{tab:app:english_rules}. 
The performance drop when using the English rules, compared to the original Chinese rules, is relatively small, indicating that the language of the grammar book has little impact on MT performance. We leave further analysis across languages in future studies.

\subsection{Induction of Descriptive Grammar Rules}

Beyond being used for the study of translation, \textsc{ZhuangBench} can be also used for the research of grammar induction. 

\paragraph{Experimental Setups} 
For each grammar rule in \textsc{ZhuangRules}, we ask LLMs to summarize a grammar rule based on its parallel sentence examples. 
We adopt two-shot ICL with Qwen-2.5-72B-Instruct,

Since directly evaluating the quality of the generated rules is challenging, we perform an extrinsic evaluation: we use the induced rules for machine translation and compare the results to those obtained using gold-standard rules.

\begin{table}[t]
\small
\centering
\setlength\tabcolsep{2.5pt}
\begin{tabular}{l|cc|cc}
\toprule
& \multicolumn{2}{c|}{\textbf{za2zh}} & \multicolumn{2}{c}{\textbf{zh2za}} \\
& \textbf{BLEU} & \textbf{chrF++} & \textbf{BLEU} & \textbf{chrF++} \\ 
\midrule
No Rule & 31.2 & 28.3 & 22.1 & 49.1\\
\midrule
Induced Rule & 45.2 & 43.4 & 45.5 & 73.4 \\ 
Gold Rule & 51.4 & 50.5 & 51.8 &  78.7 \\
\midrule
Induced R. + 2 Ex. & 67.9 & 64.2 & 73.0 & 85.5 \\
Gold R. + 2 Ex. & \textbf{70.7} & \textbf{68.0} & \textbf{80.7} & \textbf{91.0} \\
\bottomrule
\end{tabular}
\caption{Translation performance of Qwen-2.5-72B-Instruct using induced and gold rules.}
\label{tab:induction}
\end{table}

\paragraph{Results}
As shown in Table~\ref{tab:induction}, the grammar rules induced by LLMs are useful for translation. 
While the performance using induced rules is lower than using gold rules, it significantly outperforms the scenario without any rules, achieving improvements of 15.1\% and 24.3\% chrF++ for Zhuang-to-Chinese and Chinese-to-Zhuang translation, respectively. 
Additionally, pairing the induced rules with parallel examples further improves translation performance. 
We provide examples of the induced grammar rules in Table~\ref{tab:app:case_study_inductive}.

These results demonstrate the potential of using LLMs to generate grammar rules for XLR languages. 
Our pilot study focuses on clustered groups of simple parallel sentences sharing common grammatical features. Future work can explore the possibility of using LLMs to distill descriptive grammar from a larger, unordered corpus of parallel data.

\section{Case Study}
\label{app:case_study}
\label{app:case_study_translation}
We provide examples of Zhuang-to-Chinese translation from different methods in Table~\ref{tab:app:case_study_utilization}. When no additional input is provided, the model fails to translate the sentence into Zhuang, instead generating sentences with irrelevant words that do not even appear in Zhuang dictionaries, highlighting the current limitations of LLMs in supporting the Zhuang language. However, when a lexicon is provided, the translation improves significantly, though it still suffers from hallucinations and syntax errors. When rules or parallel sentence examples are provided, the model can understand the rule or mimic the examples to perform the translation, yielding better results with both lexical and syntactical accuracy. Code rules further enhance the model's ability to understand and utilize grammar rules due to the similarity between code execution and grammar rule application, and the stepwise breakdown inherent in code rules helps LLMs better comprehend complex grammar rules, also leading to improved performance. Moreover, when the test instance requires multiple rules, including all the necessary rules enables the model to generate a better translation, while using only a subset of the required rules reduces performance.

\begin{table*}
\centering
\small
\setlength{\tabcolsep}{5pt}
\begin{tabular}{p{2.0\columnwidth}r}
\toprule
\ \textbf{def check\_whether\_apply(source\_sentence, dictionary):} \\
\ \ \ \ """ \\
\ \ \ \ \zhsmall{\# 语法规则：在壮语中，单词“dwg”是系动词。在表示肯定判断的简单句中，“dwg”一般省略。然而，在表示}\\
\ \ \ \ \zhsmall{否定时，“dwg”不可省略，且否定词“mbouj”需置于系动词“dwg”之前。} \textit{(Grammar rule: In Zhuang, the word}\\
\ \ \ \ \textit{"dwg" is a copula. In affirmative simple sentences, "dwg" is usually omitted. However, in negative sentences, "dwg"}\\
\ \ \ \ \textit{cannot be omitted, and the negation word "mbouj" must be placed before the copula "dwg".)} \\
\ \\
\ \ \ \ \zhsmall{\#\# 检查是否需要应用此规则将汉语翻译为壮语的步骤如下：} \textit{(The steps to check whether this rule needs to be}\\
\ \ \ \ \textit{applied in the translation are as follows:)} \\
\ \ \ \ \zhsmall{1. 判断句子是否为判断句} \textit{(1. Determine if the sentence is a judgment sentence.)} \\
\ \ \ \ \zhsmall{2. 若是判断句，返回 True 表示需要应用该规则；否则，返回 False} \textit{(2. If the sentence is a judgment sentence, return}\\
\ \ \ \ \textit{True indicating the rule should be applied, otherwise return False.)} \\
\ \ \ \ """\\ \\
\ \ \ \ \# \zhsmall{1. 判断句子是否为判断句} \textit{(1. Determine if the sentence is a judgment sentence.)} \\
\ \ \ \ if is\_judgment\_sentence(source\_sentence): \\
\ \ \ \ \ \ \ \ \# \zhsmall{2. 若是判断句，返回 True 表示需要应用该规则} \textit{(2. If the sentence is a judgment sentence, return True.)} \\
\ \ \ \ \ \ \ \ return True \\
\ \ \ \ else: \# \zhsmall{否则，返回 False} \textit{(Otherwise, return False)} \\
\ \ \ \ \ \ \ \ return False \\
\bottomrule
\end{tabular}
\caption{An example of the code rule for rule retrieval. The \textit{text in italics} are the English translations. The code rules consist of two parts: (1) a concise comment outlining the steps for checking whether this rule needs to be applied, and (2) a pseudo-code function simulating the rule checking process.}
\label{tab:app:code_grammar_rule_retrieval}
\end{table*}

\begin{table*}
\centering
\small
\setlength{\tabcolsep}{5pt}
\begin{tabular}{p{2.0\columnwidth}r}
\toprule
\ \textbf{def apply\_rule(source\_sentence, dictionary):} \\
\ \ \ \ """ \\
\ \ \ \ \zhsmall{\# 语法规则：在壮语中，单词“dwg”是系动词。在表示肯定判断的简单句中，“dwg”一般省略。然而，在表示}\\
\ \ \ \ \zhsmall{否定时，“dwg”不可省略，且否定词“mbouj”需置于系动词“dwg”之前。} \textit{(Grammar rule: In Zhuang, the word}\\
\ \ \ \ \textit{"dwg" is a copula. In affirmative simple sentences, "dwg" is usually omitted. However, in negative sentences, "dwg"}\\
\ \ \ \ \textit{cannot be omitted, and the negation word "mbouj" must be placed before the copula "dwg".)} \\
\ \\
\ \ \ \ \zhsmall{\#\# 应用此规则将汉语翻译为壮语的步骤如下：} \textit{(The steps for applying this rule to translate Chinese into Zhuang are}\\
\ \ \ \ \textit{as follows:)} \\
\ \ \ \ \zhsmall{1. 判断句子是否为否定句} \textit{(1. Determine if the sentence is negative.)} \\
\ \ \ \ \zhsmall{2. 提取主语、谓语和宾语} \textit{(2. Extract the subject, predicate, and object.)} \\
\ \ \ \ \zhsmall{3. 翻译主语、谓语和宾语} \textit{(3. Translate the subject, predicate, and object.)} \\
\ \ \ \ \zhsmall{4. 根据句子类型（肯定或否定）调整系动词“dwg”和否定词“mbouj”的使用} \textit{(4. Adjust the use of copula "dwg" and}\\
\ \ \ \ \textit{negation word "mbouj" based on sentence type, affirmative or negative.)} \\
\ \ \ \ \zhsmall{5. 拼接成完整的句子} \textit{(5. Combine to form the complete sentence.)} \\
\ \ \ \ """\\ \\

\ \ \ \ \# \zhsmall{1. 判断句子是否为否定句} \textit{(1. Determine if the sentence is negative.)} \\
\ \ \ \ is\_negative = "\zhsmall{不是}" in source\_sentence \\
\ \ \ \ \# \zhsmall{2. 提取主语、谓语和宾语} \textit{(2. Extract the subject, predicate, and object.)} \\
\ \ \ \ subject, predicate, object\_ = extract\_subject\_predicate\_object(source\_sentence) \\
\ \ \ \ \# \zhsmall{3. 翻译主语、谓语和宾语} \textit{(3. Translate the subject, predicate, and object.)}\\
\ \ \ \ subject\_in\_zhuang = translate(subject, dictionary) \\
\ \ \ \ predicate\_in\_zhuang = translate(predicate, dictionary) \\
\ \ \ \ object\_in\_zhuang = translate(object\_, dictionary) \\
\ \ \ \ \# \zhsmall{4. 根据句子类型调整系动词和否定词} \textit{(4. Adjust the use of copula "dwg" and negation word "mbouj" based on}\\
\ \ \ \ \textit{sentence type, affirmative or negative.)}\\
\ \ \ \ if is\_negative: \# \zhsmall{否定句，使用“mbouj dwg”} \textit{(Negative sentence, use “mbouj dwg”)} \\
\ \ \ \ \ \ \ translated\_sentence = f"{subject\_in\_zhuang} mbouj dwg {predicate\_in\_zhuang} {object\_in\_zhuang}" \\
\ \ \ \ else: \# \zhsmall{肯定句，省略“dwg”} \textit{(Affirmative sentence, omit “dwg”)} \\
\ \ \ \ \ \ \ translated\_sentence = f"{subject\_in\_zhuang} {predicate\_in\_zhuang} {object\_in\_zhuang}" \\
\ \ \ \ \# \zhsmall{5. 返回完整的翻译句子} \textit{(5. Return the complete sentence.)}\\
\ \ \ \ return translated\_sentence \\
\bottomrule
\end{tabular}
\caption{An example of the code rule for rule application. The \textit{text in italics} are the English translations. The code rules consist of two parts: (1) a concise comment outlining the translation steps, and (2) a pseudo-code function simulating the translation process.}
\label{tab:app:code_grammar_rule_utilization}
\end{table*}

\begin{table*} 
\centering 
\small 
\setlength{\tabcolsep}{5pt} 
\begin{tabular}{p{2.0\columnwidth}r} 
\toprule 
\zhsmall{\# 壮语是中国的一门少数民族语言。你是一名语言学家，以下是一条壮语语法规则的相关信息。给你一些需要翻译为壮语的汉语短语或句子，请根据该语法规则的内容，逐一检查翻译过程中是否需要使用该规则。你的回答中只应包含是否需要使用该规则的判断（“是”或“否”），不包含任何其他额外信息。} \textit{(Zhuang is a minority language in China. You are a linguist, and the following is a grammar rule for the Zhuang language. You are given some Chinese phrases or sentences to translate into Zhuang; please check whether this rule applies in the translation process. Your answer should only contain whether the rule needs to be applied ("yes" or "no"), and no other additional information.)} \\ \\

\zhsmall{\#\# 语法规则：} \textit{(Grammar Rule:)} \\

\zhsmall{\#\#\# 语法规则描述：在壮语中，单词“dwg”是系动词。在表示肯定判断的简单句中，“dwg”一般省略。然而，在表示否定时，“dwg”不可省略，且否定词“mbouj”需置于系动词“dwg”之前。} \textit{(Description of the Grammar Rule: In Zhuang, the word "dwg" is a copula. In affirmative simple sentences, "dwg" is usually omitted. However, in negative sentences, "dwg" cannot be omitted, and the negation word "mbouj" must be placed before the copula "dwg".)} \\ \\

\zhsmall{\#\# 请检查下面的汉语短语或句子，判断是否需要使用该语法规则进行翻译：} \textit{(Please check the following Chinese phrase or sentence and determine whether the grammar rule needs to be applied in the translation:)} \\ \\

\zhsmall{\#\#\# 汉语短语或句子：他是我的父亲。} \textit{(Chinese phrase or sentence: He is my father.)} \\

\zhsmall{\#\#\# 在上面的短语或句子中，汉语词语“父亲”在壮语中的翻译是“daxboh”；汉语词语“他”在壮语中的翻译是“de”；汉语词语“我”在壮语中的翻译是“gou”。} \textit{(In the above phrase or sentence, the Chinese word "father" is translated into Zhuang as "daxboh"; "he" is translated as "de"; and "I" is translated as "gou.")} \\

\bottomrule 
\end{tabular} 
\caption{Prompt template for the \textsc{Rule-by-Rule} retrieval method. The \textit{italicized text} represents the English translations. When applying code rules, we directly replace the rule in the \textit{Description of the Grammar Rule} with the code-format one.}
\label{tab:app:prompt_retrieval} 
\end{table*}

\begin{table*}
\centering
\small
\setlength{\tabcolsep}{5pt}
\begin{tabular}{p{2.0\columnwidth}r}
\toprule
\zhsmall{\# 壮语是中国的一门少数民族语言。你是一名语言学家，请根据给出的信息将汉语短语或句子翻译成壮语。你的回答应该只包含翻译结果，不要包含任何其他额外信息。} \textit{(Zhuang is a minority language in China. You are a linguist, please translate the given Chinese phrases or sentences into Zhuang. Your answer should only include the translation, without any additional information.)} \\ \\
\zhsmall{\# 以下是一条关于壮语的语法规则：} \textit{(Below is a grammar rule for Zhuang:)} \\

\zhsmall{\#\# 语法规则：在壮语中，单词“dwg”是系动词。在表示肯定判断的简单句中，“dwg”一般省略。然而，在表示否定时，“dwg”不可省略，且否定词“mbouj”需置于系动词“dwg”之前。} \textit{(Description of the Grammar Rule: In Zhuang, the word "dwg" is a copula. In affirmative simple sentences, "dwg" is usually omitted. However, in negative sentences, "dwg" cannot be omitted, and the negation word "mbouj" must be placed before the copula "dwg".)} \\ \\

\zhsmall{\# 以下为该规则的一些例句及其IGT(Interlinear Glossed Text)，可以帮助你完成翻译：} \textit{(Here are some example sentences with their IGT (Interlinear Glossed Text), which can help you in the translation process:)} \\ \\

\zhsmall{\#\# 例句1：} \textit{(Example 1:)} \\
\zhsmall{字典为：\texttt{\{"de": "他", "daxboh": "父亲", "gou": "我"\}}} \textit{(Dictionary: \{"de": "he", "daxboh": "father", "gou": "I"\})} \\
\zhsmall{壮语：De mbouj dwg daxboh gou.} \\
\zhsmall{IGT：\ 3sg NEG \ COP \ 父亲 \ 1sg} \\
\zhsmall{汉语：他不是我的父亲。} \textit{(Chinese: He is not my father.)} \\ \\

\zhsmall{\#\# 例句2：} \textit{(Example 2:)} \\
\zhsmall{字典为：\texttt{\{"daxmeh": "妈", "gou": "我", "vunz": "人", "laj mbanj": "乡下"\}}} \textit{(Dictionary: \{"daxmeh": "mother", "gou": "I", "vunz": "person", "laj mbanj": "rural area"\})} \\
\zhsmall{壮语：Daxmeh gou vunz laj mbanj.} \\
\zhsmall{IGT：母亲 \ 1sg \ 人 \ 乡下} \\
\zhsmall{汉语：我妈是乡下人。} \textit{(Chinese: My mother is from the countryside.)} \\ \\

\zhsmall{\#\# 请首先写出下面的壮语短语或句子的IGT，然后将其翻译成汉语：} \textit{(Please first write the IGT of the following Zhuang phrase or sentence, and then translate it into Chinese:)} \\ \\

\zhsmall{字典为：\texttt{\{"de": "他", "daxboh": "父亲", "gou": "我"\}}} \textit{(Dictionary: \{"de": "he", "daxboh": "father", "gou": "I"\})} \\
\zhsmall{壮语：De daxboh gou.} \\
\zhsmall{\#\# 所以，该壮语短语或句子的IGT和汉语翻译分别是：} \textit{(So, the IGT and Chinese translation for the Zhuang phrase or sentence are:)} \\
\bottomrule
\end{tabular}
\caption{Prompt template for grammar rule application with 2 parallel examples and IGTs. The \textit{italicized text} represents the English translations. When applying code rules, we directly replace the rule in the \textit{Description of the Grammar Rule} with the code-format one.}
\label{tab:app:prompt_utilization}
\end{table*}


\begin{table*}[h]
\setlength\tabcolsep{3pt}
\centering
\begin{small}
\begin{tabular}{p{0.5\columnwidth}p{1.5\columnwidth}r}
\toprule
\multicolumn{2}{c}{\textit{Case 1 (Single Rule)}} \\
\midrule
Grammar Rule: & \zhsmall{壮语中表达“先做某事”用“…… gonq”，壮语中也有汉语借词sien“先”，需要将两者结合表达一个共同的意思。} \textit{(In Zhuang, the expression “to do something first” is formed with “…… gonq”; Zhuang also has the Chinese loanword sien meaning “first,” which should be combined with gonq to express the meaning.)} \\
Input: & \zhsmall{让她先讲一讲。} \textit{(Let her speak first.)} \\
Lexicon: & \zhsmall{\{"让": "hawj", "她": "de", "讲": "gangj"\}} \textit{({"let": "hawj", "she": "de", "speak": "gangj"})} \\
Gold: & Hawj de sien gangj gonq. \\ \\
No Rule (w/o Lexicon): & Gou baez boux goz daen laj. \\
No Rule (w/ Lexicon): & Hawj de gangj. \\
Parallel Examples: & Hawj de sien gangj yijz gangj gonq. \\
Textual Rule: & Hawj de sien gonq gangj. \\
Code Rule: & Hawj de sien gangj gonq. \\
Textual Rule + 2 Examples: & Hawj de sien gangj gonq. \\
Code Rule + 2 Examples: & Hawj de sien gangj gonq. \\
\midrule
\multicolumn{2}{c}{\textit{Case 2 (Multiple Rules)}} \\
\midrule
\multirow{2}{*}{Grammar Rule:} & \zhsmall{(1). 修饰语作中心语的修饰或限制时, 壮语中心语在前, 修饰语于后。} \textit{(When modifiers function as modifiers or restrictors of the head, the head appears first, followed by the modifier in Zhuang.)} \\
 & \zhsmall{(2). “yaep ndeu” 意为“一会儿，一下”，在口语中使用频率较高，用来表示时间的短暂性，需要放在句尾。} \textit{(“yaep ndeu” means “a while” or “a moment,” commonly used in spoken language to indicate the brevity of time, and should appear at the end of the sentence.)} \\
Input: & \zhsmall{我看一会儿这本书。} \textit{(I’ll read this book for a while.)} \\
Lexicon: & \zhsmall{\{"我": "gou", "看": "yawj", "这": "neix", "本": "bonj", "书": "saw"\}} \textit{({"I": "gou", "read": "yawj", "this": "neix", "measure word of books": "bonj", "book": "saw"})} \\
Gold: & Gou yawj bonj saw neix yaep ndeu. \\ \\
No Rule (w/o Lexicon): & Gou baeq boux gaj goek. \\
No Rule (w/ Lexicon): & Gou yawj neix bonj saw haeuq. \\
Textual Rule 1 Only: & Gou yawj saw neix bonj. \\
Textual Rule 2 Only: & Gou yawj neix bonj saw yaep ndeu. \\
Both Textual Rules: & Gou yawj neix bonj saw yaep ndeu. \\
Combined Code Rule: & Gou yawj saw neix bonj yaep ndeu. \\
\bottomrule
\end{tabular}
\end{small}
\caption{Two examples of Zhuang-to-Chinese translation from different methods. The \textit{text in italics} are the English translations.}
\label{tab:app:case_study_utilization}
\end{table*}

\begin{table*}[h]
\setlength\tabcolsep{3pt}
\centering
\begin{small}
\begin{tabular}{p{0.5\columnwidth}p{1.5\columnwidth}r}
\toprule
\multicolumn{2}{c}{\textit{Case 1}} \\
\midrule
Gold Grammar Rule: & \zhsmall{“yied …… yied ……”表示随着时间的推移或事物的发展，数量或程度逐渐增加或减少，类似于汉语中的“越来越……”。} (\textit{"yied ...... yied ......" indicates a gradual increase or decrease in quantity or degree over time or with development, similar to the Chinese "more and more...".)} \\
Input Parallel Example 1: & \zhsmall{汉语：我们越走就越靠近他家了。} (\textit{Chinese: The farther we walk, the closer we get to his house.}) \\
 & Zhuang: Raeuz yied byaij yied dep laeng de lo. \\
 & \zhsmall{字典为：\{"我们": "raeuz", "走": "byaij", "靠近": "dep", "家": "laeng", "他": "de"\}} (\textit{Dictionary: \{"we": "raeuz", "walk": "byaij", "approach": "dep", "house": "laeng", "his": "de"\}}) \\
Input Parallel Example 2: & \zhsmall{汉语：夏天到了，白天越来越长了。} (\textit{Chinese: Summer has arrived, and the days become longer.}) \\
 & Zhuang: Seizhah daengz lo, doengxngoenz yied daeuj yied raez. \\
 & \zhsmall{字典为：\{"夏天": "seizhah", "到": "daengz", "了": "lo", "白天": "doengxngoenz", "来": "daeuj", "长": "raez"\}} (\textit{Dictionary: \{"summer": "seizhah", "arrive": "daengz", "completed": "lo", "daytime": "doengxngoenz", "become": "daeuj", "long": "raez"\}}) \\
Induced Grammar Rule: & \zhsmall{在壮语中，当需要表达某个状态或动作的程度逐渐加深或变化时，可以在两个相关的动词或形容词之间使用“yied...yied...”结构，以强调这种渐进的变化过程。} (\textit{In Zhuang, when expressing a gradual increase or change in a state or action, the "yied...yied..." structure can be used between two related verbs or adjectives to emphasize this progressive change.}) \\
\midrule
\multicolumn{2}{c}{\textit{Case 2}} \\
\midrule
Gold Grammar Rule: & \zhsmall{“dox”+一般动词，表示动作行为是由甲乙双方相互进行的。} (\textit{"dox" + general verbs indicate that the action is performed by both parties in interaction.}) \\
Input Parallel Example 1: & \zhsmall{汉语：对骂} (\textit{Chinese: To curse at each other.}) \\
 & Zhuang: doxndaq \\
 & \zhsmall{字典为：\{"骂": "ndaq"\}} (\textit{Dictionary: \{"curse": "ndaq"\}}) \\
Input Parallel Example 2: & \zhsmall{汉语：互助} (\textit{Chinese: Mutual assistance.}) \\
 & Zhuang: doxbang \\
 & \zhsmall{字典为：\{"帮": "bang"\}} (\textit{Dictionary: \{"help": "bang"\}}) \\
Induced Grammar Rule: & \zhsmall{在壮语中，为了表达两个主体之间的互动或相互作用的行为，可以在相关动词前添加前缀“dox”，以构成表示互动或相互作用的复合动词。} (\textit{In Zhuang, to express interactive or reciprocal actions between two subjects, the prefix "dox" can be added to relevant verbs to form compound verbs indicating interaction.}) \\
\midrule
\multicolumn{2}{c}{\textit{Case 3}} \\
\midrule
Gold Grammar Rule: & \zhsmall{动词后边一般能在后面带上时态助词gvaq“过”，表示经历时态。但壮汉语在结构上略有不同, 壮语的gvaq置于述宾词组之后，汉语的“过”则紧贴着动词之后。} (\textit{Verbs in Zhuang can typically be followed by the tense marker "gvaq" to indicate the experiential past. However, Zhuang and Chinese differ in structure: Zhuang places "gvaq" \red{after the verb-object phrase}, while in Chinese, "\zhsmall{过}" follows immediately after the verb.}) \\
Input Parallel Example 1: & \zhsmall{汉语：我看过这本书。} (\textit{Chinese: I have read this book.}) \\
 & Zhuang: Gou yawj bonj saw neix gvaq. \\
 & \zhsmall{字典为：\{"我": "gou", "看": "yawj", "本": "bonj", "书": "saw", "这": "neix"\}} (\textit{Dictionary: \{"I": "gou", "read": "yawj", "measure word of book": "bonj", "book": "saw", "this": "neix"\}}) \\
Input Parallel Example 2: & \zhsmall{汉语：他到过北京。} (\textit{Chinese: He has been to Beijing.}) \\
 & Zhuang: De bae daengz Bwzgingh gvaq. \\
 & \zhsmall{字典为：\{"他": "de", "去": "bae", "到": "daengz", "北京": "Bwzgingh"\}} (\textit{Dictionary: \{"he": "de", "go": "bae", "arrive": "daengz", "Beijing": "Bwzgingh"\}}) \\
Induced Grammar Rule: & \zhsmall{在壮语中，表示过去经历的助词“gvaq”通常置于句子的末尾，用于表达动作已经发生或完成。} (\textit{In Zhuang, the experiential marker "gvaq" is usually \red{placed at the end of the sentence} to indicate that the action has already occurred or been completed.}) \\
\bottomrule
\end{tabular}
\end{small}
\caption{Examples of grammar rule inducted by Qwen-2.5-72B-Instruct. The \textit{text in italics} are the English translations.}
\label{tab:app:case_study_inductive}
\end{table*}


\end{document}